\documentclass[dvipsnames,sigconf,backref=page]{acmart}
\AtBeginDocument{%
  }

\copyrightyear{2025}
\acmYear{2025}
\setcopyright{cc}
\setcctype{by}
\acmConference[GECCO '25]{Genetic and Evolutionary Computation Conference}{July 14--18, 2025}{Malaga, Spain}
\acmBooktitle{Genetic and Evolutionary Computation Conference (GECCO '25), July 14--18, 2025, Malaga, Spain}
\acmDOI{10.1145/3712256.3726314}
\acmISBN{979-8-4007-1465-8/2025/07}

\usepackage{algorithm}
\usepackage{algorithmic}
\usepackage{xcolor}
\usepackage[capitalise]{cleveref}
\usepackage{booktabs}
\usepackage{multirow}
\usepackage{makecell}
\usepackage{mathtools}
\usepackage{subcaption}
\usepackage{enumitem}
\usepackage[skip=1pt]{caption}

\begin{document}

\title{Overcoming Deceptiveness in Fitness Optimization with Unsupervised Quality-Diversity}

\author{Lisa Coiffard}
\email{lisa.coiffard21@imperial.ac.uk}
\orcid{0009-0009-0557-9318}
\affiliation{%
    \institution{Imperial College London}
    \city{London}
    \country{United Kingdom}
}

\author{Paul Templier}
\email{p.templier@imperial.ac.uk}
\orcid{0000-0001-8919-8639}
\affiliation{%
    \institution{Imperial College London}
    \city{London}
    \country{United Kingdom}
}

\author{Antoine Cully}
\email{a.cully@imperial.ac.uk}
\orcid{0000-0002-3190-7073}
\affiliation{%
    \institution{Imperial College London}
    \city{London}
    \country{United Kingdom}
}

\renewcommand{\shortauthors}{Coiffard et al.}

\begin{abstract}
    Policy optimization seeks the best solution to a control problem according to an objective or fitness function, serving as a fundamental field of engineering and research with applications in robotics. Traditional optimization methods like reinforcement learning and evolutionary algorithms struggle with deceptive fitness landscapes, where following immediate improvements leads to suboptimal solutions. Quality-diversity (QD) algorithms offer a promising approach by maintaining diverse intermediate solutions as stepping stones for escaping local optima. However, QD algorithms require domain expertise to define hand-crafted features, limiting their applicability where characterizing solution diversity remains unclear. In this paper, we show that unsupervised QD algorithms - specifically the AURORA framework, which learns features from sensory data - efficiently solve deceptive optimization problems without domain expertise. By enhancing AURORA with contrastive learning and periodic extinction events, we propose AURORA-XCon, which outperforms all traditional optimization baselines and matches, in some cases even improving by up to 34\%, the best QD baseline with domain-specific hand-crafted features. This work establishes a novel application of unsupervised QD algorithms, shifting their focus from discovering novel solutions toward traditional optimization and expanding their potential to domains where defining feature spaces poses challenges.
\end{abstract}


\begin{teaserfigure}
    \centering
    \includegraphics[width=\textwidth]{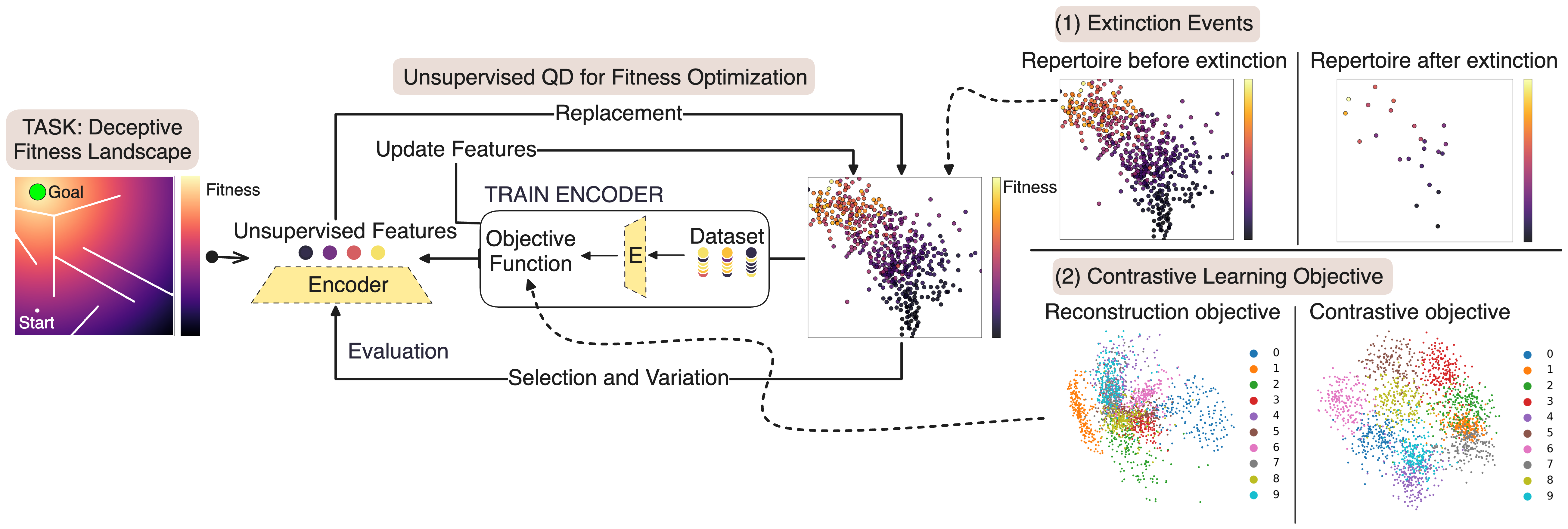}
    \caption{Unsupervised QD overcomes deceptiveness in fitness optimization tasks. We enhance this framework with (1) extinction events and (2) a contrastive learning (CL) objective. We illustrate the effect of CL by projecting digits of the MNIST dataset~\citep{deng_mnist_2012} onto learnt latent spaces trained with (left) mean-square error loss and (right) triplet loss.}
    \Description{Teaser fig}
    \label{fig:teaser}
\end{teaserfigure}


\maketitle

\section{Introduction}\label{sec:intro}

Optimization is a fundamental field of engineering and research with applications across numerous domains. From maximizing the yield of chemical reactors to optimizing the forward velocity of robots and minimizing fluctuations in wind turbine operations, optimization problems permeate modern engineering challenges. While many optimization problems can be effectively solved using fitness optimization methods such as reinforcement learning (RL) and evolutionary algorithms (EAs), we focus here on a particularly challenging class of problems characterized by deceptive fitness landscapes. In these problems, following the path of best immediate fitness does not necessarily lead to optimal solutions, but instead to an inescapable local optimum~\citep{lehman_abandoning_2011}.

Consider a robot navigating the maze presented in \cref{fig:teaser} (left): the shortest path to the goal might be blocked by walls, making strategies that initially move away from the target necessary to find the optimal solution. Lehman and Stanley~\citep{lehman_abandoning_2011} demonstrated that, in such deceptive landscapes, traditional fitness optimization approaches struggle and  become trapped in local optima, even with sophisticated exploration strategies. Instead, they propose novelty search (NS)~\citep{lehman_abandoning_2011} as a radical departure from fitness optimization by completely abandoning the objective function to reward novelty. 

NS quantifies solution novelty through a distance metric in feature space, also known as the behaviour descriptor space or characterization space~\citep{cully_quality_2018}. In maze navigation tasks, for example, the feature space typically captures the robot's terminal position, where novelty is computed as the Euclidean distance between a candidate solution's final position and those of existing solutions. This approach enables exploration of the solution space by discovering stepping stones toward optimal solutions. In this example, NS might first discover how to reach different regions of the maze before ultimately finding a path to the target, thus avoiding local optima that might trap traditional optimization methods.

Quality-diversity (QD) optimization emerged as a synthesis between NS's powerful exploration capabilities and traditional fitness optimization~\citep{pugh_confronting_2015, cully_quality_2018} to produce large collections of diverse and high-performing solutions. QD methods, such as the multi-dimensional archive of phenotyping elites (MAP-Elites) algorithm~\citep{mouret_illuminating_2015}, build on NS's ability to overcome deceptive landscapes while maintaining pressure toward optimal solutions. These methods have proven valuable across various domains, from damage-resilient locomotion by pre-computing diverse walking gaits~\citep{cully_robots_2015}, to object manipulation with varied grasp strategies~\citep{huber_quality_2025}. While originally developed for generating diverse solutions sets~\citep{mouret_illuminating_2015, templier_quality_2024, nilsson_policy_2021, pierrot_diversity_2022, faldor_synergizing_2024}, QD's ability to overcome deceptiveness makes it particularly valuable for traditional optimization problems.

Current QD algorithms rely on domain expertise to define features, which presents several key challenges. First, domain experts may not always be available or may lack sufficient understanding to define meaningful features. Second, even when expertise is available, hand-coded features might not capture task-relevant characteristics. For example, while foot contact patterns work well as features for robotic locomotion~\citep{cully_robots_2015}, determining appropriate features for manipulation tasks is far more challenging, as the interaction dynamics and task-relevant behaviours are less easily characterized~\citep{huber_quality_2025}. These limitations have restricted QD's applicability in domains where feature characterization remains difficult.

Recent work has developed unsupervised QD methods that automatically learn features from sensory inputs using dimensionality reduction~\citep{liapis_transforming_2013, paolo_unsupervised_2020, grillotti_unsupervised_2022}. These approaches eliminate the need for hand-crafted features by projecting high-dimensional trajectory data into a lower-dimensional latent space (the feature space), with a neural network encoder. \citet{liapis_transforming_2013} and  \citet{paolo_unsupervised_2020} build on NS for pure exploration, while autonomous robots realizing their abilities (AURORA)~\citep{grillotti_unsupervised_2022} extends this concept to QD optimization.

In this work, we introduce a novel application of the unsupervised QD algorithm AURORA to tackle deceptive fitness-based optimization problems. While AURORA was originally designed for QD optimization—collecting diverse and high-performing solutions across the feature space—we repurpose it specifically for maximizing objective functions. By leveraging AURORA's unsupervised feature learning, we maintain the benefits of QD's exploration without requiring hand-crafted features.

However, AURORA faces two limitations in fitness optimization settings. Its reconstruction-based feature learning does not enforce any structure on the latent space. Similar solutions might scatter across the latent space and minor variations can occupy disproportionately large areas --- a phenomenon we demonstrate through a toy experiment on the MNIST dataset (see \cref{fig:teaser}, bottom right). Additionally, online learning creates an initialization bias, whereby older solutions disproportionately influence the learnt feature space, typically around suboptimal behaviours. We address these challenges through AURORA-XCon, which enhances the base algorithm with periodic extinction events and a contrastive learning (CL) objective.

Our primary contributions include:
\begin{itemize}[noitemsep,topsep=1pt]
    \item A novel application and comprehensive evaluation of unsupervised QD methods in fitness optimization.
    \item AURORA-X: An extension incorporating periodic extinction events that promotes evolvable solutions~\citep{lehman_enhancing_2015} while mitigating initialization bias through population reduction.
    \item AURORA-Con: A CL objective that explicitly organizes the feature space according to solution performance, clustering trajectories with similar fitnesses together.
\end{itemize}

Our experimental results show that unsupervised QD algorithms solve deceptive optimization problems without the need for domain expertise in feature design. Specifically, our proposed method AURORA-XCon outperforms all traditional optimization baselines and performs comparably, in some cases even improving by up to 34\%, the best QD baseline with hand-crafted features. Our findings establish unsupervised QD as a powerful approach for optimization, expanding beyond its prior applications in discovering diverse and high-performing solutions. This opens new possibilities in domains where defining feature spaces is challenging. Our code is available at \url{https://github.com/LisaCoiffard/aurora-xcon/}.

\section{Background and Related Work}\label{sec:background}



\subsection{Quality-Diversity}\label{sec:b_qd}

In deceptive optimization problems, such as the maze environment introduced by \citet{lehman_abandoning_2011}, traditional optimization algorithms like GAs and RL typically converge to suboptimal solutions. To overcome this limitation, \citet{lehman_abandoning_2011} developed NS, which pursues behavioural novelty rather than fitness optimization. NS maps solutions into a $d$-dimensional feature space that characterizes their behaviour and maintains an archive of previously encountered solutions. The algorithm selects individuals based on their novelty score, computed as a weighted sum of distances to the $k$-nearest neighbours in feature space~\citep{doncieux_novelty_2019}, thereby promoting exploration of the feature space.

Building on the success of NS, QD algorithms emerged to combine novelty with fitness optimization~\citep{pugh_confronting_2015, cully_quality_2018, cully_behavioral_2013}. The MAP-Elites algorithm~\citep{mouret_illuminating_2015} retains the evolutionary loop of selection, variation, and replacement, but enhances it with an explicit diversity preservation mechanism. It maintains a structured repertoire parametrised by a discretized feature space, where each cell stores the highest-performing solution associated with its corresponding feature centroid. MAP-Elites forms the basis for many state-of-the-art QD optimization algorithms~\citep{nilsson_policy_2021, grillotti_unsupervised_2022, vassiliades_using_2018, faldor_synergizing_2024, pierrot_diversity_2022, fontaine_covariance_2020, fontaine_differentiable_2021}. Among these, policy gradients assisted MAP-Elites (PGA-MAP-Elites)~\citep{nilsson_policy_2021} enhances efficiency in large search spaces by incorporating a gradient-based variation operator. This approach trains critic networks with the twin delayed deep deterministic policy gradients algorithm (TD3)~\citep{fujimoto_addressing_2018} to estimate fitness gradients and guide the search towards high-performing solutions.

Quality with just enough diversity (JEDi~\citep{templier_quality_2024}) shares our objective of leveraging diversity to discover stepping stones toward optimal solutions, rather than pursuing QD optimization. JEDi extends MAP-Elites by incorporating a Gaussian process that maps the relationship between features and fitness, enabling targeted exploration of promising regions in the feature space. While both approaches employ QD algorithms for fitness optimization, JEDi requires hand-coded features, whereas our method learns these features automatically without the need for domain-specific definitions.

\subsection{Unsupervised Quality-Diversity}\label{sec:b_uqd}

Traditional QD algorithms rely on domain-specific feature spaces, limiting their applicability where appropriate features are unclear. Unsupervised QD algorithms address this limitation by learning features directly from sensory data using dimensionality reduction. This approach, pioneered by \citet{liapis_transforming_2013} and \citet{paolo_unsupervised_2020} for NS, was extended to QD optimization in AURORA~\citep{grillotti_unsupervised_2022}. AURORA integrates unsupervised feature learning with MAP-Elites to discover diverse and high-performing solutions without domain expertise.

The AURORA algorithm performs QD optimization by characterizing the feature space as a latent space, learnt through an auto-encoder neural network. Parent solutions are selected uniformly at random from an unstructured repertoire~\citep{cully_quality_2018, faldor_toward_2024}, rather than from a grid as in MAP-Elites, to accommodate the unbounded nature of the learnt feature space. Evaluated offspring state trajectories, generated through genetic variation, are encoded into this latent space and compared to existing solutions. The replacement step in AURORA relies on nearest-neighbour comparisons in the latent space to determine whether a new solution should replace an existing one. In the literature, various mechanisms have been proposed to regulate diversity in such unstructured repertoires, including volume-adaptive thresholds (AURORA-VAT) and container size control (AURORA-CSC), which dynamically adjust the distance threshold for replacement~\citep{grillotti_unsupervised_2022}.

The encoder is periodically retrained using state trajectories of evaluated solutions, with the objective of minimizing the mean-squared error (MSE) between each input trajectory and its \textit{reconstructed} output. As the algorithm converges, these training phases become less frequent. Each update of the latent space triggers a re-encoding of all repertoire solutions, ensuring that the feature representation remains consistent with the updated model.

Nevertheless, learning meaningful feature spaces for exploration remains a key challenge for AURORA. Recent work has proposed various approaches to address this limitation. RUDA~\citep{grillotti_relevance-guided_2022} incorporates task-relevance by maintaining a buffer of task-relevant solutions, though this requires a downstream task definition that may not exist. Both MC-AURORA~\citep{cazenille_ensemble_2021} and HOLMES~\citep{etcheverry_hierarchically_2020} tackle initialization bias—where early solutions overly influence feature learning—by maintaining diverse feature space representations. MC-AURORA maintains multiple repertoires with distinct features, while HOLMES employs a hierarchical architecture. However, these methods focus primarily on QD optimization rather than pure fitness optimization, which is our focus.

\begin{figure*}[htb]
  \centering
  \includegraphics[width=\textwidth]{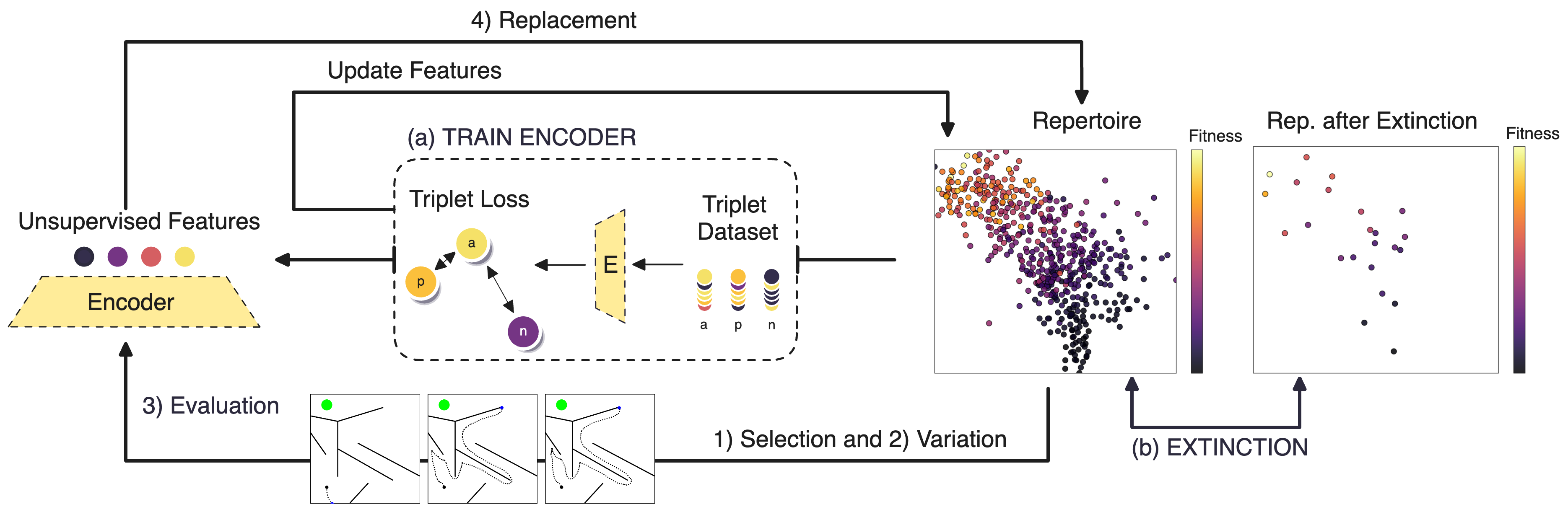}
  \caption{Overview of the AURORA-XCon algorithm with two key contributions: (a) encoder training with a contrastive objective (the triplet loss) and (b) periodic extinction events.}
  \Description{Method Diagram}
  \label{fig:method}
\end{figure*}

\subsection{Continual Learning}\label{sec:related_work_plasticity}

The online learning process of the AURORA algorithm introduces a fundamental challenge associated with non-stationary distributions: initialization bias. Early in the evolutionary process, solutions typically cluster in suboptimal regions of the fitness landscape. As the encoder learns to differentiate between these solutions, it shapes the feature space around potentially similarly low-quality trajectories. This bias stems from a broader challenge in continual learning, where neural networks struggle to balance preserving existing knowledge with integrating new information~\citep{dohare_loss_2024}. When AURORA subsequently discovers truly novel solutions, the encoder tends to project them into confined regions of the feature space, effectively diminishing meaningful differences and compromising the algorithm's ability to maintain stepping stones toward optimal solutions.

Previous work has addressed the plasticity challenge through different approaches. For instance, \citet{dohare_loss_2024} employed continual back-propagation with neural unit re-initialization to maintain network adaptability. We tackle this challenge with periodic extinction events, building on insights from \citet{lehman_enhancing_2015}, who demonstrated that stochastic population reduction enhances evolvability in divergent search strategies. Extinctions serve a dual purpose in this work: they mitigate initialization bias by regularly refreshing the encoder's training data, while simultaneously promoting the emergence of solutions capable of rapid adaptation to new niches.

\section{Method}\label{sec:method}

Building on AURORA's foundation in unsupervised feature learning for QD algorithms, we introduce two targeted enhancements designed specifically for optimization tasks, illustrated in \cref{fig:method}. First, we replace AURORA's reconstruction objective with contrastive learning (CL) to create a more structured feature space organization. This modification enforces clusters over similar fitnesses, while maintaining appropriate separation between trajectories that lead to distinct fitness values. Second, we implement periodic extinction events that systematically refresh the population, addressing the initialization bias inherent in online feature learning and promoting the emergence of highly adaptable solutions.

The complete algorithm, presented in \cref{alg:aurora-xcon}, combines these enhancements in a modular framework that yields four variants: 
\begin{itemize}[noitemsep,topsep=1pt]
    \item AURORA: the original algorithm
    \item AURORA-X: with extinction events
    \item AURORA-Con: with CL objective
    \item AURORA-XCon: the complete framework.
\end{itemize}

\subsection{Contrastive Learning of Features}\label{sec:m_triplet}

AURORA-XCon replaces the traditional reconstruction objective with CL through a triplet loss~\citep{schroff_facenet_2015} (\cref{fig:method} (a)). This modification enables the encoder to organize solutions in the feature space according to their relative performance, despite fitness information not being directly present in the input trajectories.

To illustrate the impact of this architectural choice, we conduct a toy experiment using the MNIST dataset~\citep{deng_mnist_2012}, visualized in \cref{fig:teaser} (bottom right). We train two auto-encoders on identical data: one with a traditional reconstruction objective (MSE loss) and another with our contrastive objective (triplet loss). The two-dimensional latent embeddings reveal the differing organizational principles of the two objectives. The reconstruction objective, while effectively preserving information needed to reconstruct the digits, produces a latent space where similar digits (e.g., 3, 5, and 8) significantly overlap. Moreover, it allocates disproportionate regions to easily reconstructed digits (e.g., 0s and 1s) and compresses more complex digits (e.g., 5s, 6s, and 8s) into smaller areas. This uneven distribution can prove problematic for QD algorithms, where the latent space organization directly influences which solutions are maintained in the repertoire. In contrast, the contrastive objective creates a structured latent space with clear clustering and uniform area allocation across digit classes. As part of AURORA, the encoder learns to extract performance-relevant features from state trajectories, even though fitness values are only provided during the loss computation. This results in a more efficient optimization process, where the feature space organization actively supports the discovery of increasingly better solutions rather than merely maintaining diversity.

To train the encoder with the triplet loss, we pre-process the repertoire's solutions into groups of three: an anchor solution $a$, a positive example $p$ with similar fitness, and a negative example $n$ with dissimilar fitness. We iterate through the repertoire and for each solution, we select it as the anchor and form triplets. The triplet's positive and negative examples are selected as the solution with closest fitness ($p$) and furthest fitness ($n$), from a randomly sampled pair over the rest of the repertoire. The loss function encourages the encoder to satisfy:

\begin{equation}
   d(a,p) + m < d(a,n)
\end{equation}

where $d(x_i,x_j)$ represents the Euclidean distance between encoded features of solutions $x_i$ and $x_j$, and $m$ is a margin parameter enforcing minimum separation. In other words, the distance between solutions with similar fitness ($d(a,p)$) should be smaller than the distance to solutions with dissimilar fitness ($d(a,n)$) by at least margin $m$. This objective is formalized as:

\begin{equation}
   \mathcal{L}_{\text{triplet}}= \sum^N_i \big[\max(d(a,p) - d(a,n) + m, 0)\big], \quad m > 0
\end{equation}

The margin parameter $m$ controls the strength of separation between similar and dissimilar solutions in the feature space. Rather than treating the margin as a fixed hyperparameter, we dynamically set $m = d_{min}$, where $d_{min}$ is the minimum distance between any pair of solutions based on the current state of the unstructured repertoire. This adaptive margin ensures appropriate separation between fitness-based clusters as the repertoire's population evolves.

\subsection{Extinction Events}\label{sec:m_extinction}

Building on insights from continual learning research, we address initialization bias through periodic extinction events (\cref{fig:method} (b)). While previous approaches like continual back-propagation~\citep{dohare_loss_2024} focus on maintaining network plasticity, our extinction mechanism takes a population-level approach and is inspired by \citet{lehman_enhancing_2015}'s work on evolvability in divergent search.

Every $T_E$ iterations, the algorithm randomly preserves only $k\%$ of solutions (set to $k=5\%$ in our experiments) while maintaining the single highest-fitness solution through elitism. This mechanism serves two complementary purposes, one addressing the continual learning challenge and another to enhance exploration in QD. First, extinction directly counters initialization bias by regularly refreshing the encoder's training distribution, preventing early solutions from permanently dominating feature learning. The stochastic nature of the preservation, combined with elitism, ensures that while the feature space can adapt to new solutions, progress toward optimization is maintained. Second, the periodic population bottlenecks create evolutionary pressure favouring solutions with high evolvability~\citep{lehman_enhancing_2015} --- those capable of quickly radiating to fill multiple niches after extinction. In our optimization context, extinction events could promote the discovery of stepping stone solutions, as their lineages demonstrate the ability to efficiently explore promising regions of the search space.

\begin{algorithm}[t]
\caption{AURORA-XCon algorithm}\label{alg:aurora-xcon}
\begin{algorithmic}[1]
\REQUIRE Batch size $b$, encoder update iterations $U_1, \ldots, U_N$, extinction period $T_E$, extinction proportion $\epsilon$
\STATE Initialize population $X$ and encoder $\xi$ with $b$ random solutions
\FOR{$i = 1$ to $I$}
    \STATE $x_1, \ldots, x_b \gets \textsc{Selection}(X)$
    \STATE $\hat{x}_1, \ldots, \hat{x}_b \gets \textsc{Variation}(x_1, \ldots, x_b)$
    \FOR{each $\hat{x}_1, \ldots, \hat{x}_b$}
        \STATE $f \gets F(\hat{x})$
        \STATE $\phi \gets \xi(s_0, \ldots, s_T)$
        \STATE $X \gets \textsc{Replacement}(\hat{x})$
    \ENDFOR
    \IF{$i \in \{U_1, \ldots, U_N\}$}
        \STATE \colorbox{yellow}{$X_{\textsc{Triplet}} \gets \textsc{Pre-process}(X)$}
        \STATE \colorbox{yellow}{$\xi \gets \textsc{Train\_Triplet}(\xi, X_{\textsc{Triplet}})$}
        \FOR{each $x \in X$}
            \STATE $\phi \gets \xi(s_0, \ldots, s_T)$
        \ENDFOR
    \ENDIF
    \IF{$i \mod T_E = 0$}
        \STATE \colorbox{yellow}{$X \gets \textsc{Extinction}(X, \epsilon)$}
    \ENDIF
\ENDFOR
\end{algorithmic}
\end{algorithm}

\section{Experimental Setup}\label{sec:experimental_setup}

\begin{figure}[ht]
  \centering
  \includegraphics[width=\linewidth]{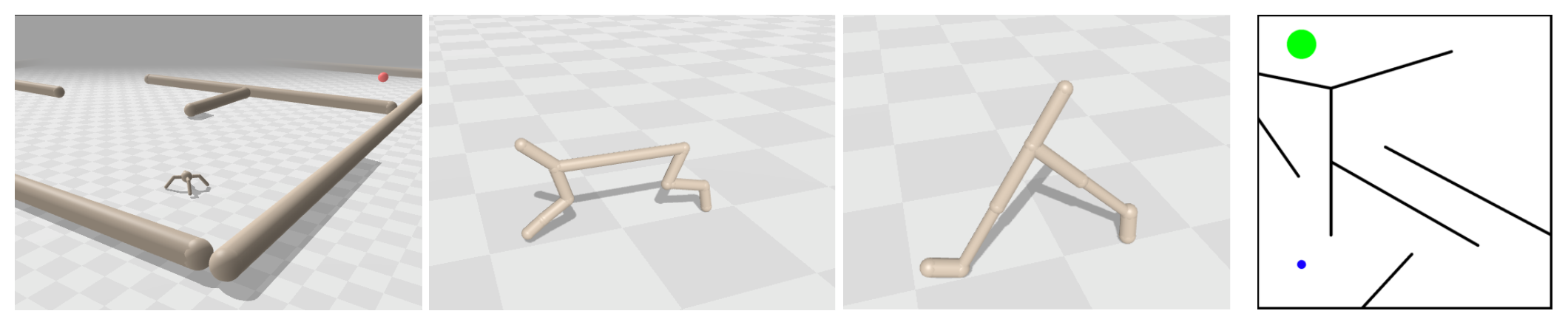}
  \caption{Overview of the environments used in our experiments. From left to right: \textsc{AntMaze}, \textsc{HalfCheetah}, \textsc{Walker} and \textsc{Kheperax} standard maze.}
  \label{fig:environments}
  \Description{Environments}
\end{figure}

\begin{table}[ht] 
    \caption{Evaluation Tasks}
    \label{table:robotics_envs}
    \begin{center}
    \begin{tabular}{@{}rcccc@{}}
        \toprule
        & \textsc{AntMaze} & \textsc{H.Cheetah} & \textsc{Walker} & \textsc{Khep.} \\
        \midrule
        $|\mathcal{S}|$ & 29 & 18 & 17 & 5 \\
        $|\mathcal{A}|$ & 8 & 6 & 6 & 2 \\
        Ep. len. & 1000 & 1000 & 1000 & 200 \\
        Policy size & [128, 128] & [128, 128] & [128, 128] & [5] \\
        Params. & 20,352 & 18,944 & 18,816 & 30 \\
        Traj. size & [100, 29] & [100, 18] & [100, 17] & [50, 5] \\
        $\phi$ & $xy$-pos. & feet contact & feet contact & $xy$-pos. \\
        \bottomrule
    \end{tabular}  
    \end{center}
\end{table}

\subsection{Environments and Tasks}\label{sec:exp_envs}

We evaluate AURORA-XCon across four robotics control tasks, visualized in \cref{fig:environments}, from \textsc{Brax}~\citep{freeman_brax_2021} and \textsc{Kheperax}~\citep{grillotti_kheperax_2023} benchmarks. The \textsc{Brax} benchmark provides three tasks: \textsc{AntMaze}, a three-dimensional maze navigation challenge, and two two-dimensional locomotion tasks, \textsc{HalfCheetah} and \textsc{Walker}. The fourth task comes from \textsc{Kheperax} and presents a two-dimensional maze navigation problem. The navigation tasks require robots to minimize their final distance to a goal, marked by a red ball and green circle respectively. These environments are inherently deceptive, as the optimal path requires moving away from the target. The locomotion tasks challenge robots to maximize their forward velocity, presenting an unbounded optimization problem. Both benchmarks are built on JAX~\citep{bradbury_jax_2018}, enabling efficient parallel evaluation of our experiments.

\cref{table:robotics_envs} summarizes the key properties of these environments. For baseline comparisons using hand-coded features, we adopt established features, $\phi$, from the QD literature: foot contact patterns for locomotion tasks and terminal goal distance for navigation challenges, all mapped to two-dimensional feature spaces. The table also details the dimensionality of state trajectories (traj. size) provided to AURORA's encoder, highlighting the need for dimensionality reduction in creating meaningful feature representations.

\subsection{Baselines}\label{sec:exp_baselines}

We evaluate AURORA-XCon across a comprehensive suite of baselines, detailed below.

\textbf{Vanilla GA and TD3.} To demonstrate the value of diversity maintenance, we compare against methods focused solely on performance optimization. Our implementation of a vanilla GA samples parent solutions uniformly at random, applies variation using the directional operator introduced in MAP-Elites~\citep{vassiliades_discovering_2018}, and retains the top-performing individuals within a fixed-size population. For a gradient-based comparison, we include TD3~\citep{fujimoto_addressing_2018}, a state-of-the-art reinforcement learning algorithm widely used in robotics control. Given the \textsc{Kheperax} environment's relatively small search space (\cref{table:robotics_envs} params.), we exclude TD3 from these experiments.

\textbf{MAP-Elites and PGA-MAP-Elites.} To establish an upper \\ bound on performance, we implement MAP-Elites with established domain-specific features. For the larger search spaces of \textsc{AntMaze}, \textsc{HalfCheetah}, and \textsc{Walker} (approximately 20,000 parameters, see \cref{table:robotics_envs}), we also evaluate PGA-MAP-Elites~\citep{nilsson_policy_2021}, which enhances efficiency through TD3-based gradient estimates. These methods require carefully designed features based on domain expertise, in contrast to our unsupervised methods which learn these features automatically.

\textbf{AURORA and PGA-AURORA.} We compare against AURORA, the foundation for our work and the state-of-the-art in unsupervised QD. We adopt the same selection and replacement scheme across of all AURORA-based variants presented in this work, including AURORA-X, AURORA-Con, and AURORA-XCon. We also evaluate PGA-AURORA~\citep{chalumeau_neuroevolution_2022} variants that leverage gradient estimates in the same way as PGA-MAP-Elites. 

\textbf{JEDi.} Finally, we evaluate against JEDi, which shares our goal of leveraging MAP-Elites' diversity maintenance to accelerate optimization. While both approaches use diversity to guide search toward promising solutions, JEDi, like MAP-Elites, relies on domain-specific hand-crafted features, whereas we learn these features in an unsupervised manner.

\subsection{Experimental Details}

We implement all experiments using QDax~\citep{lim_accelerated_2022}, benchmarking each algorithm for one million evaluations across 20 random seeds.
For comparison with evolutionary methods, we ensure TD3 experiences an equivalent number of environment interactions ($1 \times 10^9$ steps). However, TD3's algorithm requires a critic update at each step, resulting in substantially more training updates than PGA-based methods ($1 \times 10^9$ versus $5.9 \times 10^6$ critic updates). Due to increased computation time from training the critic, we limit TD3 experiments to either $1 \times 10^9$ steps or 72 hours. Full hyperparameters are detailed in \cref{app:hyperparams}.

We evaluate the performance of the algorithms by tracking the highest fitness value across all repertoire solutions at regular intervals. For TD3, we maintain a \textit{passive repertoire} accumulating solutions from policy rollouts at fixed intervals. Statistical significance is assessed using Wilcoxon rank-sum tests~\citep{mann_test_1947} with Holm-Bonferroni correction~\citep{holm_simple_1979}. For the \textsc{Kheperax} maze task specifically, we compute p-values based on evaluations-to-goal rather than final fitness since most methods eventually solve the maze (maximum fitness of zero).

\section{Results}\label{sec:results}

\begin{table}[htbp]
\caption{Final maximum fitness, with best fitnesses in bold.}
\label{tab:results}
\begin{center}
\begin{tabular}{@{}llr@{}}
\toprule
\textbf{Task} & \textbf{Method} & \textbf{Median (IQR)} \\
\midrule
\multirow{6}{*}{\textsc{Walker}}
    & GA & 1344.55 (139.22) \\
    & TD3 & 3275.00 (3284.16) \\
    & JEDi & 1225.81 (76.12) \\
    & PGA-MAP-Elites & 3809.49 (1638.59) \\
    & PGA-AURORA & 2532.10 (799.82) \\
    & PGA-AURORA-XCon & \textbf{5115.10 (2812.66)} \\
\midrule
\multirow{6}{*}{\textsc{HalfCheetah}}
    & GA & 4721.95 (675.33) \\
    & TD3 & 5036.92 (1725.56) \\
    & JEDi & 4977.46 (262.21) \\
    & PGA-MAP-Elites & 6712.53 (211.47) \\
    & PGA-AURORA & 6530.58 (340.50) \\
    & PGA-AURORA-XCon & \textbf{6950.68 (191.93)} \\
\midrule
\multirow{6}{*}{\textsc{AntMaze}} 
    & GA & -15.82 (0.10) \\
    & TD3 & -15.87 (0.02) \\
    & JEDi & -11.21 (11.17) \\
    & PGA-MAP-Elites & -3.83 (4.84) \\
    & PGA-AURORA & -4.52 (2.79) \\
    & PGA-AURORA-XCon & \textbf{-3.16 (3.84)} \\
\midrule
\multirow{5}{*}{\textsc{Kheperax}}
    & GA & -12.11 (0.19) \\
    & JEDi & -11.94 (12.09) \\
    & MAP-Elites & \textbf{0.00 (0.00)} \\
    & AURORA & \textbf{0.00 (0.00)} \\
    & AURORA-XCon & \textbf{0.00 (0.00)} \\
\bottomrule
\end{tabular}
\end{center}
\end{table}

\begin{figure*}[ht]
  \centering
  \includegraphics[width=\linewidth]{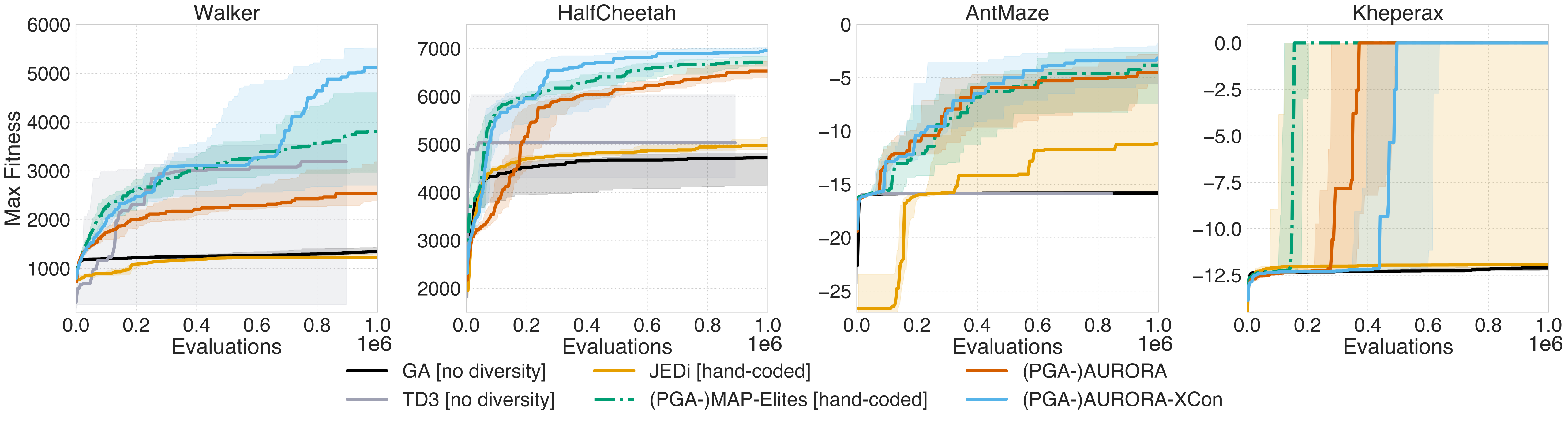}
  \caption{Maximum fitness tracked over 1 million evaluations. We show non-PGA-variants for \textsc{Kheperax} and PGA-variants for all \textsc{Brax} tasks. We plot the median (solid line) and interquartile range (IQR, shaded area).}
  \label{fig:main_results_locomotion}
  \Description{Main results: Brax}
\end{figure*}

\begin{figure*}[ht]
  \centering
  \includegraphics[width=\linewidth]{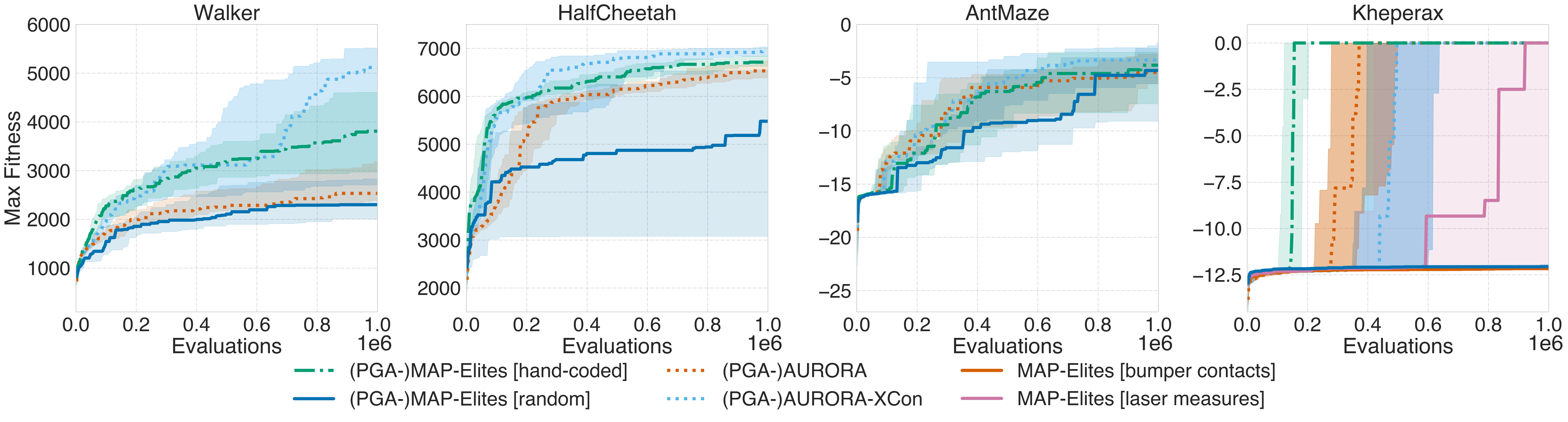}
  \caption{Maximum fitness tracked over 1 million evaluations. We show non-PGA-variants for \textsc{Kheperax} and PGA-variants for all \textsc{Brax} tasks. We plot the median (solid line) and IQR (shaded area).}
  \label{fig:desc_ablation}
  \Description{Desc ablation}
\end{figure*}

In this section, we structure our experimental results to address three research questions (\textbf{RQ1-RQ3}). 
\textbf{RQ1:} How does unsupervised QD perform on fitness optimization tasks compared to the considered baselines (\cref{sec:res_main})? 
\textbf{RQ2:} What role does feature space definition play on the effectiveness of QD algorithms for optimization (\cref{sec:res_desc_ablations})? 
\textbf{RQ3:} To what extent do CL and extinction events enhance AURORA's optimization capabilities (\cref{sec:res_ext_ablations})?

\subsection{Comparison with Baselines}
\label{sec:res_main}

In \cref{fig:main_results_locomotion} and \cref{tab:results}, we demonstrate that unsupervised QD algorithms effectively solve fitness optimization problems by leveraging diversity without requiring domain expertise. Our empirical evaluation shows that these not only outperform non-diversity-based algorithms but also match or exceed the performance of (PGA)-MAP-Elites, our upper-bound baseline that benefits from domain knowledge.

Both AURORA and AURORA-XCon significantly outperform the GA ($p < 10^{-9}$) and TD3 ($p < 10^{-7}$) --- our baselines with no diversity maintenance --- across all tasks except \textsc{Walker}, where TD3 matches AURORA's performance. While these methods plateau at suboptimal fitness values early in training, unsupervised QD exhibits sustained improvement throughout the learning process. The \textsc{Walker} task presents an interesting case where TD3 shows strong early learning, albeit with high variance across runs. Notably, TD3 was allocated 170 times more critic training steps than the QD algorithms, making it a particularly strong baseline.

In the \textsc{Brax} tasks (\textsc{AntMaze}, \textsc{Walker} and \textsc{HalfCheetah}), PGA-AURORA-XCon demonstrates strong performance, achieving comparable or better results than both its predecessor PGA-AURORA and the hand-coded feature, upper-bound baseline PGA-MAP-Elites. Our approach shows significant improvements on unbounded fitness tasks, outperforming both baselines in \textsc{HalfCheetah} ($p < 10^{-3}$) and \textsc{Walker} ($p < 10^{-4}$). These results are striking, as PGA-MAP-Elites benefits from hand-coded features, while our unsupervised approaches learn these from state trajectories.

The \textsc{Kheperax} maze navigation task presents a more nuanced picture. While both AURORA and AURORA-XCon successfully solve the maze within 500,000 evaluations, they require 1.5 (AURORA) and 3 times (AURORA-XCon) as many evaluations to so. Furthermore, AURORA-XCon takes longer than AURORA to reach the maximal fitness, suggesting that the benefits of our proposed enhancements may be task-dependent.

When compared to JEDi, which similarly leverages QD's diversity mechanism for fitness optimization, PGA-AURORA and PGA-AURORA-XCon demonstrate significantly better performance across all \textsc{Brax} tasks ($p < 10^{-3}$). In the \textsc{Kheperax} maze environment, JEDi exhibits bimodal performance with a large variance: runs either discover the goal rapidly or become permanently trapped in local optima. We note a substantial discrepancy between JEDi's performance in our locomotion tasks and the results reported in \citet{templier_quality_2024}, likely due to differences in \textsc{Brax} back-end (legacy spring vs. v1), the former of which appears more challenging for JEDi.

\subsection{Impact of Feature Space Definition}
\label{sec:res_desc_ablations}

To demonstrate that the feature space learnt by unsupervised QD methods captures information that is relevant to solving the task, we compare (PGA)-AURORA and (PGA)-AURORA-XCon to a variant of (PGA)-MAP-Elites with random features. For this, the algorithm selects random state dimensions at initialization along which to extract two-dimensional feature vectors. Feature values are extracted at evaluation from random trajectory steps and normalized using empirically determined bounds, to accommodate MAP-Elites' requirement for bounded feature spaces. Our results, presented in \cref{fig:desc_ablation}, demonstrate the benefits of the feature space learnt by (PGA)-AURORA and (PGA)-AURORA-XCon. Furthermore, they illustrate the improvement gap between random and domain-specific feature characterization for QD algorithms. 

Both (PGA)-AURORA variants demonstrate robust performance, consistently outperforming random feature selection ($p < 10^{-3}$), except in \textsc{AntMaze}. These results underscore a key advantage of unsupervised QD methods: by learning meaningful representations of state trajectories, they offer a more robust alternative to both random feature selection and potentially suboptimal hand-crafted features. 

When compared to (PGA)-MAP-Elites with hand-coded features, the former significantly outperforms our random variant on the \textsc{HalfCheetah} ($p < 10^{-3}$), \textsc{Kheperax} ($p < 10^{-7}$) and \textsc{Walker} ($p < 10^{-4}$) tasks. While the \textsc{AntMaze} environment shows a more modest improvement, the random variant shows slower convergence to high-performing solutions with a large variance across runs.

We conduct an additional experiment on the \textsc{Kheperax} maze task, comparing three distinct feature parametrisations for MAP-Elites. The first parametrisation is the hand-coded variant, which aligns directly with the optimization goal, using the robot's terminal $xy$-position in the maze. The other two derive features from sensor data and have no direct alignment with the navigation task: a two-dimensional space based on the proportion of bumper contacts and a three-dimensional space constructed from mean laser range measurements.

MAP-Elites with $xy$-position features achieves optimal performance within 200,000 evaluations, substantially outperforming both sensor-based alternatives. The laser-based features eventually enable maze completion but require approximately 900,000 evaluations ($p < 10^{-5}$), while the bumper-based features prove ineffective ($p < 10^{-8}$). This performance disparity underscores the importance of selecting features that meaningfully relate to the optimization objective.

\subsection{Impact of AURORA-XCon Enhancements}
\label{sec:res_ext_ablations}

\begin{figure}[tb]
  \centering
  \includegraphics[width=\linewidth]{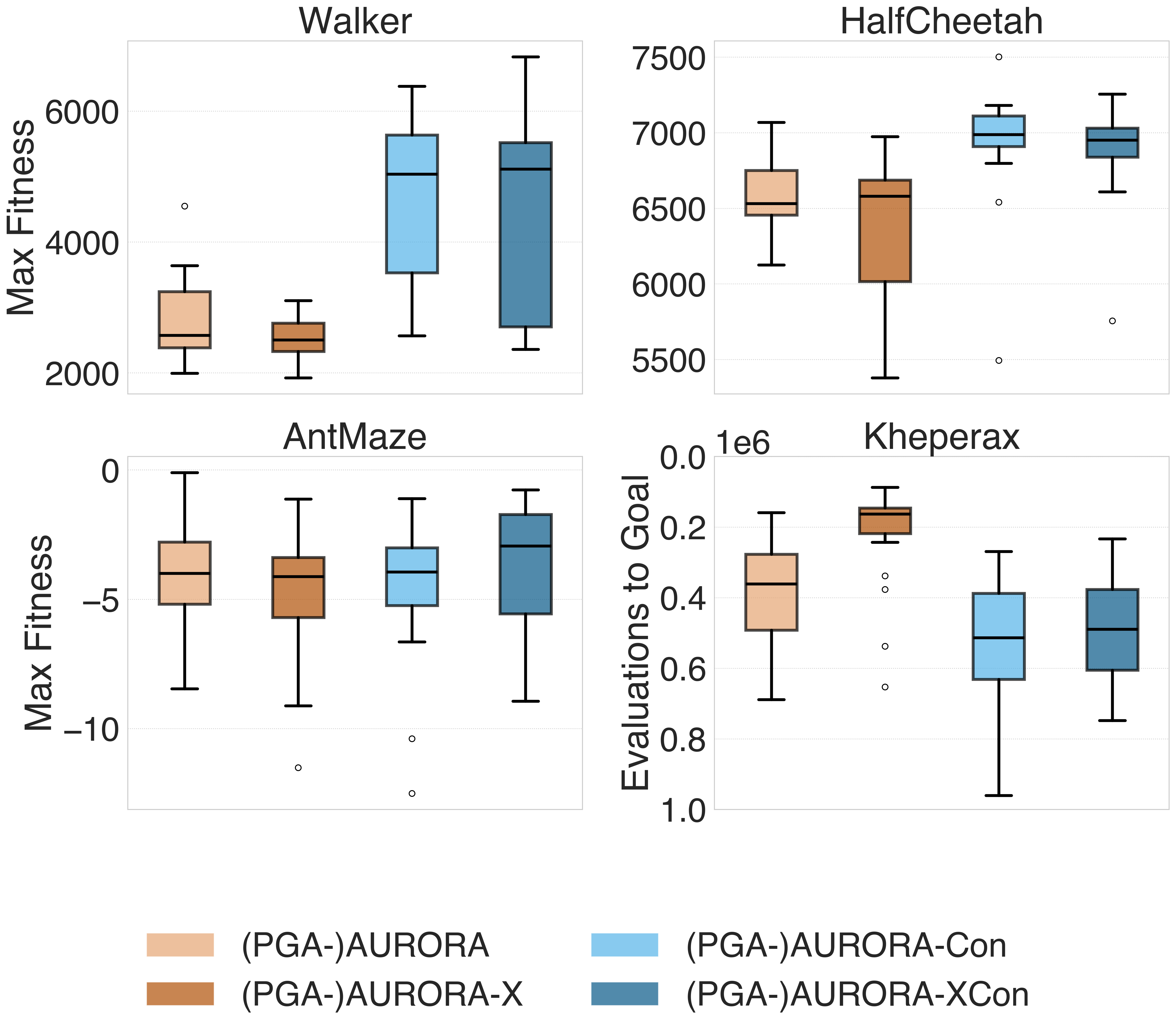}
  \caption{Final performance after 1 million evaluations, showing median and IQR. Results display evaluations-to-goal for \textsc{Kheperax} (non-PGA-variants) and maximum fitness for \textsc{Brax} tasks (PGA-variants).}
  \label{fig:ablation_locomotion}
  \Description{Brax ablation}
\end{figure}

\begin{figure}[ht]
  \centering
  \includegraphics[width=\linewidth]{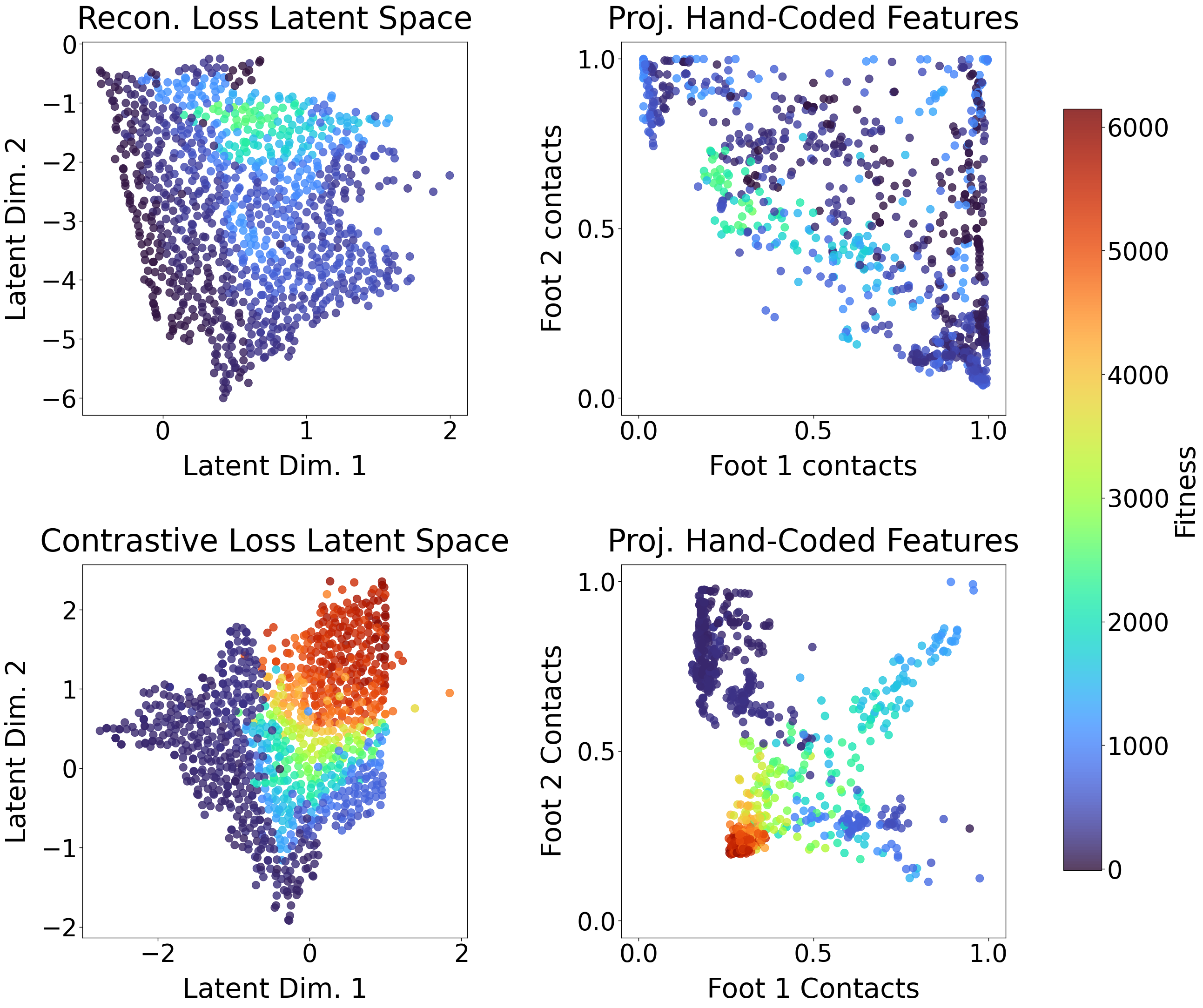}
  \caption{Comparison of the reconstruction (top) and contrastive (bottom) latent spaces learnt by AURORA's encoder and their projected features in \textsc{Walker}.}
  \label{fig:latent_comp}
  \Description{Latent comp}
\end{figure}

\cref{fig:ablation_locomotion} presents our evaluation of the individual and combined effects of our key algorithmic components. The integration of CL (PGA-AURORA-Con) yields significant performance improvements over standard PGA-AURORA across both locomotion tasks ($p < 10^{-4}$). 
\cref{fig:latent_comp} provides a qualitative comparison of the repertoires produced by AURORA (top) and AURORA-Con (bottom) at the end of 1 million evaluations on the \textsc{Walker} task. While the reconstruction objective produces broad coverage of the hand-coded feature space, it over-represents low-performing solutions. In contrast, the contrastive objective achieves a balanced representation across fitness values and concentrates exploration in high-performing regions. However, CL's performance advantage does not extend to navigation tasks, where AURORA achieves optimal fitness with fewer evaluations than AURORA-Con in the \textsc{Kheperax} environment. 

Instead, extinction events prove more valuable for navigation challenges, as illustrated in \cref{fig:ablation_locomotion}. In \textsc{AntMaze}, AURORA-XCon outperforms its non-extinction counterpart, while in \textsc{Kheperax}, both AURORA-X and AURORA-XCon achieve better sample efficiency. The improvement is especially notable for AURORA-X over AURORA in \textsc{Kheperax} ($p < 10^{-3}$). These results suggest that periodic population resets mitigate initialization bias in deceptive environments, where early on, the evolutionary process tends to accumulate similar solutions that are stuck in local optima. For locomotion tasks, extinction events show neither significant benefits nor drawbacks, with AURORA-X and AURORA-XCon performing comparably to the baseline in both \textsc{Walker} and \textsc{HalfCheetah} environments. These results highlight the task-dependent nature of our algorithmic components --- while CL consistently benefits locomotion tasks, extinction events primarily enhance navigation performance. 

\section{Conclusions and Future Work}

In this work, we show that unsupervised QD algorithms can effectively solve deceptive optimization problems without requiring domain expertise in feature design. Building on the AURORA framework, originally developed for QD optimization, we propose AURORA-XCon, which augments unsupervised feature learning with a contrastive objective and periodic extinction events. Our empirical results demonstrate that AURORA-XCon outperforms traditional optimization methods and matches or exceeds QD baselines with hand-crafted features. Ablation studies highlight the task-dependent benefits of each component: contrastive learning aids locomotion tasks, while extinction events improve navigation performance. These findings establish unsupervised QD as a powerful approach for traditional optimization problems, expanding beyond its conventional role in discovering diverse solutions and opening new possibilities in domains where defining feature spaces poses significant challenges.

These results open several avenues for future work. Unsupervised QD exhibits high variance across runs, stemming from the online learning of the feature space alongside the evolutionary process. This could be mitigated by enhancing the encoder’s capacity using continual representation learning techniques~\citep{dohare_loss_2024, rao_continual_2019}. While extinction events help counter initialization bias, their effectiveness is task-dependent, suggesting the need for task-specific tuning or more general mechanisms. Furthermore, although our method performs strongly, it converges more slowly in some cases (e.g., \textsc{Kheperax}) and incurs additional computational cost due to auto-encoder training. 
These trade-offs underscore the need to explore alternative dimensionality reduction methods that can better generalize across diverse testbeds and domains.

\begin{acks}
This research was supported in part by funding from Luffy AI Limited towards the PhD scholarship of Lisa Coiffard.
\end{acks}

\bibliographystyle{ACM-Reference-Format}
\bibliography{bib}

\newpage

\appendix

\section*{Supplementary Materials}

\section{Hyperparameters}\label{app:hyperparams}

Hyperparameters for all experiments are shown in the following sections. All our baselines with variation use the directional variation operator introduced in~\citep{vassiliades_discovering_2018} with parameter values $\sigma_1$ and $\sigma_2$ detailed in the tables below.

\subsection{GA}

\begin{table}[H]
\caption{GA hyperparameters}
\begin{tabular}{lc}
\toprule
\textsc{Parameter} & \textsc{Value} \\
\midrule
Total evaluations & $1 \times 10^{6}$ \\
Evaluation batch size & 512 \\
Policy networks & [128, 128, $|\mathcal{A}|$] \\
& ([5, $|\mathcal{A}|$] for \textsc{Khep.}) \\
\midrule
Population size  & 1024 \\
GA variation param. 1 ($\sigma_1$) & 0.005 \\
& (0.2 for \textsc{Khep.}) \\
GA variation param. 2 ($\sigma_2$) & 0.05 \\
& (0 for \textsc{Khep.}) \\
\bottomrule
\end{tabular}
\end{table}

\subsection{TD3}

\begin{table}[H]
\caption{TD3 hyperparameters}
\begin{tabular}{lc} 
\toprule
\textsc{Parameter} & \textsc{Value} \\
\midrule
Total environment steps & $1 \times 10^9$ \\
Warm-up steps & 2,000 \\
Gradient updates per step & 1 \\
Policy exploration noise & 0.1 \\
\midrule
Replay buffer size & $10^6$\\
TD3 batch size & 100\\
Critic network &  [256, 256, 1] \\
Actor network &  [128, 128, $|\mathcal{A}|$] \\
Critic learning rate & $3 \times 10^{-4}$ \\
Actor learning rate & $3 \times 10^{-4}$ \\
Discount factor & 0.99\\
Actor delay & 2\\
Target update rate & 0.005 \\
Smoothing noise var. & 0.2 \\
Smoothing noise clip & 0.5 \\
\bottomrule
\end{tabular}
\end{table}

\subsection{MAP-Elites}

\begin{table}[H]
\label{tab:me_hp}
\caption{MAP-Elites hyperparameters}
\begin{tabular}{lc} 
\toprule
\textsc{Parameter} & \textsc{Value} \\
\midrule
Total evaluations & $1 \times 10^{6}$ \\
Evaluation batch size & 512 \\
Policy networks & [128, 128, $|\mathcal{A}|$] \\
& ([5, $|\mathcal{A}|$] for \textsc{Khep.}) \\
\midrule
Num. centroids  & 1024 \\
GA variation param. 1 ($\sigma_1$) & 0.005 \\
& (0.2 for \textsc{Khep.}) \\
GA variation param. 2 ($\sigma_2$) & 0.05 \\
& (0 for \textsc{Khep.}) \\
\bottomrule
\end{tabular}
\end{table}

\subsection{JEDi}

For JEDi, we report task-specific hyperparameters, namely the number of ES generations before sampling a new set of targets (ES steps) and the $\alpha$ value, tuned to each environment in \cref{tab:jedi_hp}. We refer the reader to \citet{templier_quality_2024} for an in-depth explanation of these parameters. 

\begin{table}[H]
\centering
\caption{JEDi hyperparameters}
\label{tab:jedi_hp}
\begin{tabular}{lc} 
\toprule
\textsc{Parameter} & \textsc{Value} \\
\midrule
Total evaluations $N_e$ & $1 \times 10^{6}$ \\
Evaluation batch size $b$ & 512 \\
Policy networks & [128, 128, $|\mathcal{A}|$] \\
& ([8, $|\mathcal{A}|$] for \textsc{Khep.}) \\
Num. centroids $N_C$  & 1024 \\
\midrule
ES & LM-MA-ES \\
& (Sep-CMA-ES for \textsc{Khep.}) \\
Population per ES & 256 \\
& (16 for \textsc{Khep.}) \\
ES number & 4 \\
& (16 for \textsc{Khep.}) \\
Evaluation batch size & 1,024 \\
& (256 for \textsc{Khep.}) \\
\midrule
\textsc{Walker} \\
\hspace{3em} ES steps & 1,000 \\
\hspace{3em} $\alpha$ & 0.3 \\
\midrule
\textsc{HalfCheetah} \\
\hspace{3em} ES steps & 100 \\
\hspace{3em} $\alpha$ & 0.1 \\
\midrule
\textsc{AntMaze} \\
\hspace{3em} ES steps & 1,000 \\
\hspace{3em} $\alpha$ & 0.5 \\
\midrule
\textsc{Kheperax} \\
\hspace{3em} ES steps & 100 \\
\hspace{3em} $\alpha$ & 0.7 \\
\bottomrule
\end{tabular}
\end{table}

\newpage

\subsection{AURORA}

We implement all AURORA variants using an unstructured repertoire with a local competition mechanism inspired by dominated novelty search (DNS)~\citep{bahlous-boldi_dominated_2025}. This departs from AURORA's original distance thershold-based mechanism. Our approach ranks candidates for replacement based on their distance to and fitness dominance by other individuals in the repertoire. This promotes diversity by preserving distinct high-performing solutions while avoiding the need for sensitive threshold tuning.

For all AURORA variants, we use a single-layer LSTM auto-encoder to capture temporal dependencies in the state trajectories. The encoder's latent space dimensionality, which determines our feature space dimension, is specified in \cref{tab:aurora_hp}. The encoder is trained at linearly increasing intervals (see \textit{encoder update interval} in \cref{tab:aurora_hp}) with an early-stopping mechanism when the loss fails to improve by more than $0.0005$ for $10$ consecutive steps to prevent overfitting.

In AURORA-(X)Con variants, following \cref{sec:m_triplet}, we dynamically set the triplet margin $m$ to the minimum distance $d_{min}$ between any pair of solutions in the repertoire. This value is computed as the repertoire adapts to new individuals.

\begin{table}[H]
\caption{AURORA hyperparameters}
\label{tab:aurora_hp}
\begin{tabular}{lc} 
\toprule
\textsc{Parameter} & \textsc{Value} \\
\midrule
Total evaluations & $1 \times 10^{6}$ \\
Evaluation batch size & 512 \\
Policy networks & [128, 128, $|\mathcal{A}|$] \\
& ([5, $|\mathcal{A}|$] for \textsc{Khep.}) \\
\midrule
Repertoire max. size  & 1024 \\
GA variation param. 1 ($\sigma_1$) & 0.005 \\
& (0.2 for \textsc{Khep.}) \\
GA variation param. 2 ($\sigma_2$) & 0.05 \\
& (0 for \textsc{Khep.}) \\
\midrule
Feature space dimensionality ($h$) & 10 \\
Encoder batch size & 128 \\
Encoder learning rate & $1 \times 10^{-2}$ \\
Encoder training epochs & 200 \\ 
Encoder update interval & 10 \\
\midrule
Extinction period ($T_E$) & 50 \\
Prop. remaining after extinction ($k$) & 0.05 \\
Triplet margin ($m$) & $h \times d_{min}$ \\
\bottomrule
\end{tabular}
\end{table}

\newpage

\subsection{PGA-based Methods}

PGA-MAP-Elites and PGA-AURORA(-XCon) use the PGA-specific hyperparameters detailed in \cref{tab:pga_hp}, while maintaining their respective base algorithm parameters as specified in the previous sections.

\begin{table}[H]
\caption{PGA-* hyperparameters}
\label{tab:pga_hp}
\begin{tabular}{lc} 
\toprule
\textsc{Parameter} & \textsc{Value} \\
\midrule
Total evaluations & $1 \times 10^{6}$ \\
Evaluation batch size & 512 \\
Policy networks & [128, 128, $|\mathcal{A}|$] \\
& ([5, $|\mathcal{A}|$] for \textsc{Khep.}) \\
\midrule
Repertoire max. size  & 1024 \\
GA variation param. 1 ($\sigma_1$) & 0.005 \\
& (0.2 for \textsc{Khep.}) \\
GA variation param. 2 ($\sigma_2$) & 0.05 \\
& (0 for \textsc{Khep.}) \\
\midrule
GA mutation proportion & 0.5 \\
Critic training steps & 3000 \\
PG variation learning rate & $5 \times 10^{-3}$ \\
PG training steps & 150\\
\midrule
Replay buffer size & $10^6$\\
TD3 batch size & 100\\
Critic network &  [256, 256, 1] \\
Actor network &  [128, 128, $|\mathcal{A}|$] \\
Critic learning rate & $3 \times 10^{-4}$ \\
Actor learning rate & $3 \times 10^{-4}$ \\
Discount factor & 0.99\\
Actor delay & 2\\
Target update rate & 0.005 \\
Smoothing noise var. & 0.2 \\
Smoothing noise clip & 0.5 \\
\bottomrule
\end{tabular}
\end{table}

\end{document}